\documentclass[conference]{IEEEtran}

\newif\ifcheckliststyle
\checkliststylefalse

\usepackage{amsmath,amssymb,amsfonts}
\usepackage{graphicx}
\usepackage{booktabs}
\usepackage{multirow}
\usepackage{microtype}
\usepackage{siunitx}
\usepackage{url}
\usepackage{cite}
\usepackage{textcomp}
\usepackage{xcolor}
\usepackage{capt-of}
\usepackage{afterpage}
\usepackage{tablefootnote}

\makeatletter
\newcommand{\setcaptype}[1]{\def\@captype{#1}}
\makeatother
\newsavebox{\tempboxtwo}

\usepackage[hidelinks]{hyperref}

\def\BibTeX{{\rm B\kern-.05em{\sc i\kern-.025em b}\kern-.08em  T\kern-.1667em\lower.7ex\hbox{E}\kern-.125emX}}

\begin{document}

\title{Diagnosing Shortcut-Induced Rigidity in Continual Learning: The Einstellung Rigidity Index (ERI)}

\author{
  \IEEEauthorblockN{Kai Gu\textsuperscript{*} and Weishi Shi}
  \IEEEauthorblockA{
    Department of Computer Science and Engineering\\
    University of North Texas\\
    Denton, TX 76203, USA\\
    Email: \{kaigu@my.unt.edu, weishi.shi@unt.edu\}
  }
  \thanks{\textsuperscript{*}Corresponding author.}
}

\maketitle

\begin{abstract}
Deep neural networks frequently exploit shortcut features, defined as incidental correlations between inputs and labels without causal meaning. Shortcut features undermine robustness and reduce reliability under distribution shifts. In continual learning (CL), the consequences of shortcut exploitation can persist and intensify: weights inherited from earlier tasks bias representation reuse toward whatever features most easily satisfied prior labels, mirroring the cognitive \emph{Einstellung} effect, a phenomenon where past habits block optimal solutions. Whereas catastrophic forgetting erodes past skills, shortcut-induced rigidity throttles the acquisition of new ones. We introduce the Einstellung Rigidity Index (ERI), a compact diagnostic that disentangles genuine transfer from cue-inflated performance using three interpretable facets: (i) Adaptation Delay (AD), (ii) Performance Deficit (PD), and (iii) Relative Suboptimal Feature Reliance (SFR\_rel). 

On a two-phase CIFAR-100 CL benchmark with a deliberately spurious magenta patch in Phase~2, we evaluate Na\"ive fine-tuning (SGD), online Elastic Weight Consolidation (EWC\_on), Dark Experience Replay (DER++), Gradient Projection Memory (GPM), and Deep Generative Replay (DGR). Across these continual learning methods, we observe that CL methods reach accuracy thresholds earlier than a Scratch-T2 baseline (negative AD) but achieve slightly lower final accuracy on patched shortcut classes (positive PD). Masking the patch improves accuracy for CL methods while slightly reducing Scratch-T2, yielding negative SFR\_rel.  This pattern indicates the patch acted as a \emph{distractor} for CL models in this setting rather than a helpful shortcut.

While these findings are consistent with shortcut-induced rigidity, we caution that results may vary with backbones, datasets, and cue salience. Overall, ERI serves as a litmus test to screen tasks for rigidity risks and guide interventions: computed alongside standard CL metrics, it clarifies whether apparent adaptation proceeds for the right reasons and flags rigidity risks when suspected cues can be intervened on.
\end{abstract}

\begin{IEEEkeywords}
Continual learning; Shortcut Learning; Spurious Correlations; Einstellung Effect; Robustness; Evaluation Metrics; Representation Similarity; Convolutional Neural Networks; Deep Learning.
\end{IEEEkeywords}

\section{Introduction}

Over the past decade, research in continual learning (CL) has predominantly focused on one failure mode: catastrophic forgetting (CF). When a model learns a new task, gradients from new data can overwrite parameters that supported earlier tasks and reduce retention of prior knowledge \cite{doi:10.1073/pnas.1611835114,vandeven2024continuallearningcatastrophicforgetting}. 

A broad range of approaches—replay buffers \cite{NEURIPS2019_fa7cdfad, chaudhry2018efficient, buzzega2020darkexperiencegeneralcontinual}, parameter regularizers \cite{doi:10.1073/pnas.1611835114,pmlr-v70-zenke17a}, dynamic architectures \cite{rusu2022progressiveneuralnetworks,yoon2018lifelonglearningdynamicallyexpandable}, rehearsal schedules \cite{buzzega2020darkexperiencegeneralcontinual, Wang_2021_CVPR}, generative replay \cite{DBLP:journals/corr/ShinLKK17}, and subspace projection constraints \cite{DBLP:journals/corr/abs-2103-09762}—has been developed to slow, mask, or alleviate this erosion.
 
Yet forgetting is only one half of the stability--plasticity dilemma. The same mechanisms that protect prior representations can reduce plasticity: instead of discarding the past, the model may preferentially reuse features from earlier tasks, possibly including shortcuts that are suboptimal for the current task. This pattern is analogous to the \emph{Einstellung effect}, where prior strategies impede discovery of better ones \cite{Binz2021ReconstructingTE}. We refer to this counterpart as \emph{shortcut-induced rigidity}. Whereas CF concerns retention of prior tasks, rigidity concerns over-reliance on suboptimal features during adaptation.
 
This consideration can be viewed through a bias--variance analogy, although the mapping is heuristic. Regularization-based methods such as EWC \cite{doi:10.1073/pnas.1611835114} and Synaptic Intelligence \cite{pmlr-v70-zenke17a} constrain parameter updates to reduce forgetting, which can also reduce the degrees of freedom available for new learning and thereby entrench early-learned shortcuts \cite{Kong2022BalancingSA}. Conversely, highly plastic approaches may adapt quickly on new tasks but exhibit greater variability in the retention of past knowledge.
 
The consequences are most salient when models exploit shortcuts—incidental correlations (e.g., distinctive colors, textures, or background artifacts) that predict labels without causal meaning \cite{Geirhos_2020,Hermann2023OnTF}. In CL, shortcuts learned on earlier tasks may become entrenched through the same mechanisms that prevent forgetting. If regularizers protect parameters encoding spurious features, or if replay buffers contain many shortcut-bearing examples, adaptation on subsequent tasks can be biased toward reusing these correlations \cite{Murali2023ShortcutLT}, potentially creating a feedback loop in which anti-forgetting mechanisms also preserve shortcut reliance.
 
We introduce the \emph{Einstellung Rigidity Index} (ERI), a three-facet diagnostic designed to assess shortcut-induced rigidity in sequential learning. ERI quantifies: (i) how quickly a continual learner reaches an accuracy threshold relative to a from-scratch baseline on shortcut-bearing data (Adaptation Delay, AD); (ii) the final accuracy gap on those data (Performance Deficit, PD); and (iii) the additional reliance of the continual model on the suspected shortcut compared to the from-scratch baseline, measured via a masking intervention (Relative Sub-optimal Feature Reliance, SFR\(_{\mathrm{rel}}\)). ERI complements standard CL metrics by distinguishing genuine transfer from cue-inflated performance.
 
In this work, we articulate shortcut-induced rigidity as the adaptation-side counterpart to catastrophic forgetting and propose the Einstellung Rigidity Index (ERI) to assess it without altering training. ERI captures learning speed, final performance, and cue sensitivity via AD, PD, and SFR\(_{\mathrm{rel}}\). Using a two-phase CIFAR-100 protocol with deterministic shortcut injection and masking, we compare Na\"ive fine-tuning, EWC, Experience Replay, and related methods to a from-scratch baseline and find negative AD, small positive PD, and negative SFR\(_{\mathrm{rel}}\) on shortcut classes. This indicates faster adaptation alongside increased sensitivity to the injected cue (which behaves as a distractor in our configuration), rather than shortcut reuse. We also discuss screening and mitigation practices, and position ERI as a complement to conventional CL metrics by evaluating whether adaptation proceeds for the right reasons.

\section{Related Work}
\subsection{Shortcut learning and spurious correlations}
Shortcut learning represents a fundamental challenge in deep learning \cite{Steinmann2024navigating}. This phenomenon manifests across diverse domains: texture bias in computer vision \cite{Geirhos_2020}, background artifacts in medical imaging \cite{Hill2024TheRO,Bassi2024improving}, and device-specific signatures in audio processing \cite{shim2023constructperfectworsethancoinflipspoofing}. Contemporary detection methods include mutual information-based monitoring \cite{Adnan2022MonitoringSL,Fay2023AvoidingSB}, counterfactual interventions \cite{Robinson2022DeepLM}, and training dynamics analysis \cite{murali2023distributionshiftspuriousfeatures}. Particularly relevant to our work, Murali et al. \cite{murali2023distributionshiftspuriousfeatures} demonstrate that shortcuts correspond to features learned early in training, introducing Prediction Depth as an instance-difficulty metric. This temporal perspective informs our Adaptation Delay component in ERI. 

\subsection{Continual learning and evaluation}
Traditional continual learning evaluation relies on aggregate metrics such as average accuracy (ACC), backward transfer (BWT), and forward transfer (FWT) \cite{vandeVen2024ContinualLA}. While invaluable, these do not distinguish whether success stems from robust, causal features or from spurious cues \cite{Ashley2021DoesTA,Marconato2022CatastrophicFI, Wang2023WhereTF,Sun2024RevivingDM}. When spurious correlations are present, they can both accelerate learning (by supplying an easy signal) and impair learning (by suppressing gradients for more informative features), leading to misleading progress indicators and reduced scalability. Hammoud et al. \cite{hammoud2023rapidadaptationonlinecontinual} demonstrate that online continual learning metrics can be manipulated through spurious label correlations, proposing temporally-aware evaluation that accounts for spurious correlation effects. Similarly, domain-agnostic characterization frameworks \cite{BAKER2023274} showcase the need for evaluation protocols that can detect when apparent learning progress masks underlying brittleness. Salman et al. \cite{salman2022doesbiastransfertransfer} systematically study bias transfer in sequential learning, showing that spurious correlations are not only inherited but often amplified across tasks. Salmani \& Lewis \cite{salmani2024transferlearningbias} further demonstrate that transfer learning can introduce new biases not present in either source or target domains, supporting our hypothesis about weight inheritance creating representation rigidity. Our metric addresses these evaluation gaps by providing a diagnostic that, when suspected cues are known or can be approximated through targeted interventions, can detect shortcut-induced rigidity. In settings where cues are unknown, we outline practical proxies and their limitations in the Discussion, and we recommend treating ERI as a conservative screen rather than a definitive attribution.

\subsection{Cognitive biases in machine learning}
The Einstellung effect, where prior knowledge interferes with optimal problem-solving, has recently gained attention as a framework for understanding limitations in machine learning systems \cite{Binz2021ReconstructingTE}. This cognitive bias manifests when established solution strategies prevent the discovery of better alternatives, directly paralleling our observations in continual learning scenarios. In the context of transfer learning, Székely et al. \cite{10.1162/opmi_a_00158} demonstrate that transfer learning systematically reshapes inductive biases, showing how prior tasks influence subsequent learning patterns. This finding supports our hypothesis that weight inheritance in continual learning creates rigidity by biasing representation search toward previously successful but potentially suboptimal features. The connection between cognitive biases and machine learning limitations provides theoretical grounding for our approach. 

\section{Methodology}

\subsection{Rationale for Two Phases}
Our goal is to diagnose whether prior training induces \emph{rigidity} that
steers adaptation toward an easy but suboptimal cue. A two-phase
continual-learning (CL) protocol cleanly separates:
\begin{enumerate}
  \item Phase~1 (T1): acquisition of broadly useful features in the absence of
  the cue; and
  \item Phase~2 (T2): adaptation to new labels in the presence of a suspected
  shortcut cue.
\end{enumerate}
This separation enables attribution of faster learning or higher apparent
accuracy in Phase~2 either to genuine transfer or to shortcut reuse, which ERI
disentangles along three facets.

\subsection{Phase Definitions and Evaluation}
\textbf{Phase~1 (T1):} Learn generally useful, cue-free representations.

\textbf{Phase~2 (T2):} Introduce new superclasses; a designated subset is
augmented with a deterministic shortcut cue. We evaluate:
\begin{itemize}
  \item with the shortcut cue present; and
  \item under a masking intervention that removes the cue.
\end{itemize}

\subsection{Formal Definition}
We consider a two-phase CL setting. Let \(\mathcal{Y}_{\mathrm{T2}}\) be the Phase~2 label set and \(\mathcal{C}_{\text{patch}}\subseteq\mathcal{Y}_{\mathrm{T2}}\) the subset of Phase~2 shortcut superclasses whose images are augmented with a spurious patch during Phase~2. All ERI quantities are computed on \(\mathcal{C}_{\text{patch}}\).

Let \(\mathcal{D}_{\mathrm{T2}}^{\text{test,patch}}\) denote the Phase~2 test set restricted to \(\mathcal{C}_{\text{patch}}\) with the patch present, and \(\mathcal{D}_{\mathrm{T2}}^{\text{test,mask}}\) the same underlying images with the patch removed by a deterministic masking operator. For any model \(M\), define \(\mathrm{Acc}(M,\mathcal{D})\) as top-1 accuracy macro-averaged over \(\mathcal{C}_{\text{patch}}\).

We compare a Scratch-T2 model \(M_S\) trained from random initialization on Phase~2 only and a continual model \(M_{CL}\) obtained by continuing training from Phase~1. 

Let \(e\in\mathbb{N}_0\) index effective Phase~2 epochs: the number of optimizer updates that consume Phase~2 samples divided by the size of the Phase~2 training set. For methods with replay, this normalizes for extra updates due to memory sampling. Let \(\mathcal{A}_M(e)\) denote the patched-set top-1 accuracy after \(e\) effective epochs (evaluated at the checkpoint saved after epoch \(e\)). Concretely, if a training step mixes real T2 and replayed samples, we count only the real T2 portion toward \(e\); generated or replayed samples do not increase \(e\), ensuring comparability across methods.

\subsection{Adaptation Delay (AD)}
Fix an \emph{accuracy threshold} \(\tau\in(0,1)\), defined as the target macro-averaged top-1 accuracy on \(\mathcal{D}_{\mathrm{T2}}^{\text{test,patch}}\) over \(\mathcal{C}_{\text{patch}}\). Intuitively, the first effective epoch at which a model’s accuracy curve crosses \(\tau\) is its time-to-threshold. The effective epochs to reach \(\tau\) are

\begin{align}
E_S(\tau) &= \min\{e:\mathcal{A}_S(e)\ge\tau\}, \label{eq:es_tau}\\
E_{CL}(\tau) &= \min\{e:\mathcal{A}_{CL}(e)\ge\tau\}. \label{eq:ecl_tau}
\end{align}

If \(\tau\) is never reached within the budget, \(E(\tau)\) is right-censored and AD is reported as undefined. We set \(\tau{=}0.6\) (chance is \(1/|\mathcal{C}_{\text{patch}}|\); with two classes this is \(0.5\)), giving a 10-point margin above chance. To reduce threshold-crossing noise, we apply a moving-average smoothing of width \(w{=}3\) over \(\mathcal{A}_M(e)\) prior to computing \(E(\tau)\).
The Adaptation Delay is

\begin{equation}
\mathrm{AD} = E_{CL}(\tau)-E_S(\tau).
\label{eq:ad}
\end{equation}

Substantially negative \(\mathrm{AD}\) indicates the continual model reaches \(\tau\) faster. When accompanied by elevated shortcut reliance (Section~\ref{ssec:sfr}), this pattern is consistent with shortcut-accelerated learning rather than improved semantic transfer. 

\subsection{Performance Deficit (PD)}
Let \(\mathcal{A}_M^*\) denote the final patched-set accuracy of model \(M\) under the selection rule of best Phase~2 validation accuracy (the same rule for all methods). Define

\begin{equation}
\mathrm{PD} = \mathcal{A}_S^*-\mathcal{A}_{CL}^*.
\label{eq:pd}
\end{equation}

Positive \(\mathrm{PD}\) means the continual model underperforms the scratch baseline on patched data; negative \(\mathrm{PD}\) indicates an apparent advantage for the continual model that may be cue-inflated if paired with high shortcut reliance.

\subsection{Relative Suboptimal Feature Reliance (SFR\(_{\mathrm{rel}}\))}
\label{ssec:sfr}

For any \(M\), define the masking delta

\begin{equation}
\Delta_M = \mathrm{Acc}\!\left(M,\mathcal{D}_{\mathrm{T2}}^{\text{test,patch}}\right) -\mathrm{Acc}\!\left(M,\mathcal{D}_{\mathrm{T2}}^{\text{test,mask}}\right).
\label{eq:delta}
\end{equation}

Here, \(\Delta_M>0\) means performance drops when the shortcut is removed (shortcut reliance), whereas \(\Delta_M<0\) means the patch is ignored or harmful. The relative reliance is

\begin{equation}
\mathrm{SFR}_{\mathrm{rel}} = \Delta_{CL}-\Delta_S.
\label{eq:sfr_rel}
\end{equation}

Positive \(\mathrm{SFR}_{\mathrm{rel}}\) indicates that the continual process amplified dependence on the shortcut compared to the scratch baseline.

When the suspected cue is helpful for the baseline (i.e., \(\Delta_S>0\)), \(\mathrm{SFR}_{\mathrm{rel}}>0\) indicates greater shortcut reliance by the continual model. When the cue is harmful (i.e., \(\Delta_S\le 0\)), the sign has a different interpretation: \(\mathrm{SFR}_{\mathrm{rel}}>0\) means the continual model is less hurt by the cue than Scratch\_T2, not that it relies more on the cue. For cue-harmful regimes, we optionally offer a magnitude-based sensitivity,
\begin{equation}
\mathrm{CSR}_{\mathrm{rel}} = |\Delta_{CL}| - |\Delta_S|
\label{eq:CSR}
\end{equation}

where \(\mathrm{CSR}_{\mathrm{rel}}>0\) indicates the continual model is more sensitive to the cue than Scratch\_T2, regardless of sign.

\subsection{The ERI}
We define the Einstellung Rigidity Index as the ordered triplet
\begin{equation}
\mathrm{ERI} = \big(\mathrm{AD},\,\mathrm{PD},\,\mathrm{SFR}_{\mathrm{rel}}\big).
\label{eq:eri}
\end{equation}

Interpreting the three facets jointly is essential.

\begin{itemize}
 \item Red-flag pattern (likely shortcut-induced rigidity when the cue is helpful, \(\Delta_S>0\)): \(\mathrm{AD}\ll 0\), \(\mathrm{PD}\le 0\), and \(\mathrm{SFR}_{\mathrm{rel}}>0\).
 \item Benign transfer (no elevated shortcut reliance):
    \(\mathrm{AD}\approx 0\) or \(>0\),
    \(\mathrm{PD}\ge 0\),
    \(\mathrm{SFR}_{\mathrm{rel}}\le 0\).
 \item Cue-harmful regime (\(\Delta_S\le 0\)): interpret \(\mathrm{SFR}_{\mathrm{rel}}\) as relative harm. Optionally use \(\mathrm{CSR}_{\mathrm{rel}}>0\) to flag greater cue sensitivity by the continual model.
 \item Ambiguous cases should warrant additional probes (e.g., representational drift via CKA, calibration under masking, counterfactual patch placement).
\end{itemize}

For decision support, one may optionally declare a high-rigidity flag if, for user-chosen margins \(\delta_{\mathrm{AD}},\delta_{\mathrm{PD}},\delta_{\mathrm{SFR}}>0\), \(\mathrm{AD}\le -\delta_{\mathrm{AD}}\), \(\mathrm{PD}\le -\delta_{\mathrm{PD}}\), and \(\mathrm{SFR}_{\mathrm{rel}}\ge \delta_{\mathrm{SFR}}\).

\section{Experiment Design}
\subsection{Dataset, tasks, and shortcut injection}
We construct a two-phase CL benchmark on CIFAR-100~\cite{Krizhevsky2009LearningML}.

\textbf{Phase~1 (T1).} A ResNet-18 backbone~\cite{DBLP:journals/corr/HeZRS15} is trained on 8 CIFAR-100 \emph{superclasses} using standard augmentations (random crop and horizontal flip). This stage provides a semantic foundation intended to encourage robust feature learning.

\textbf{Phase~2 (T2).} The model is continually trained on 4 new superclasses. Of these, 2 are designated as \emph{shortcut superclasses} ($SC$): every image from those superclasses is augmented with a fixed \(4\times4\) magenta patch (RGB 255,\,0,\,255) in the top-left corner. The other 2 are \emph{non-shortcut superclasses} ($NSC$), which receive no patch.
This design enables a within-phase contrast between shortcut-bearing and purely
semantic classes.

\begin{figure}[htbp]
 \centering
 \includegraphics[width=\columnwidth]{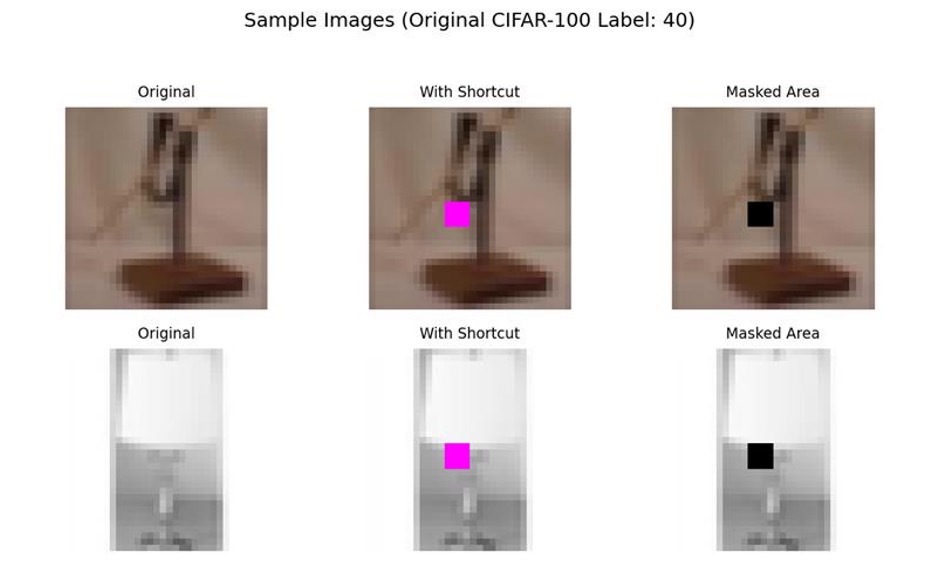}
 \caption{Shortcut injection and masking protocol. Left: original CIFAR-100
 images. Middle: shortcut added (magenta \(4\times4\) patch in the top-left).
 Right: masked evaluation replaces the patch with a uniform black square. Both
 color and grayscale examples are shown to emphasize cue salience under
 photometric variation.}
 \label{fig:samples}
\end{figure}

The patch is applied after spatial augmentations and before normalization, so
its location is consistent across training and evaluation. At test time, we
additionally evaluate a masking intervention: the same \(4\times4\)
region is overwritten with a black square (RGB 0,\,0,\,0). The masked set is a
deterministic transform of the patched set, enabling paired comparisons.
Figure~\ref{fig:samples} shows examples of original, shortcut-injected, and
masked images, highlighting the high-contrast, non-semantic nature of the cue.

\subsection{Data augmentation}
\begin{figure}[htbp]
 \centering
 \includegraphics[width=\columnwidth]{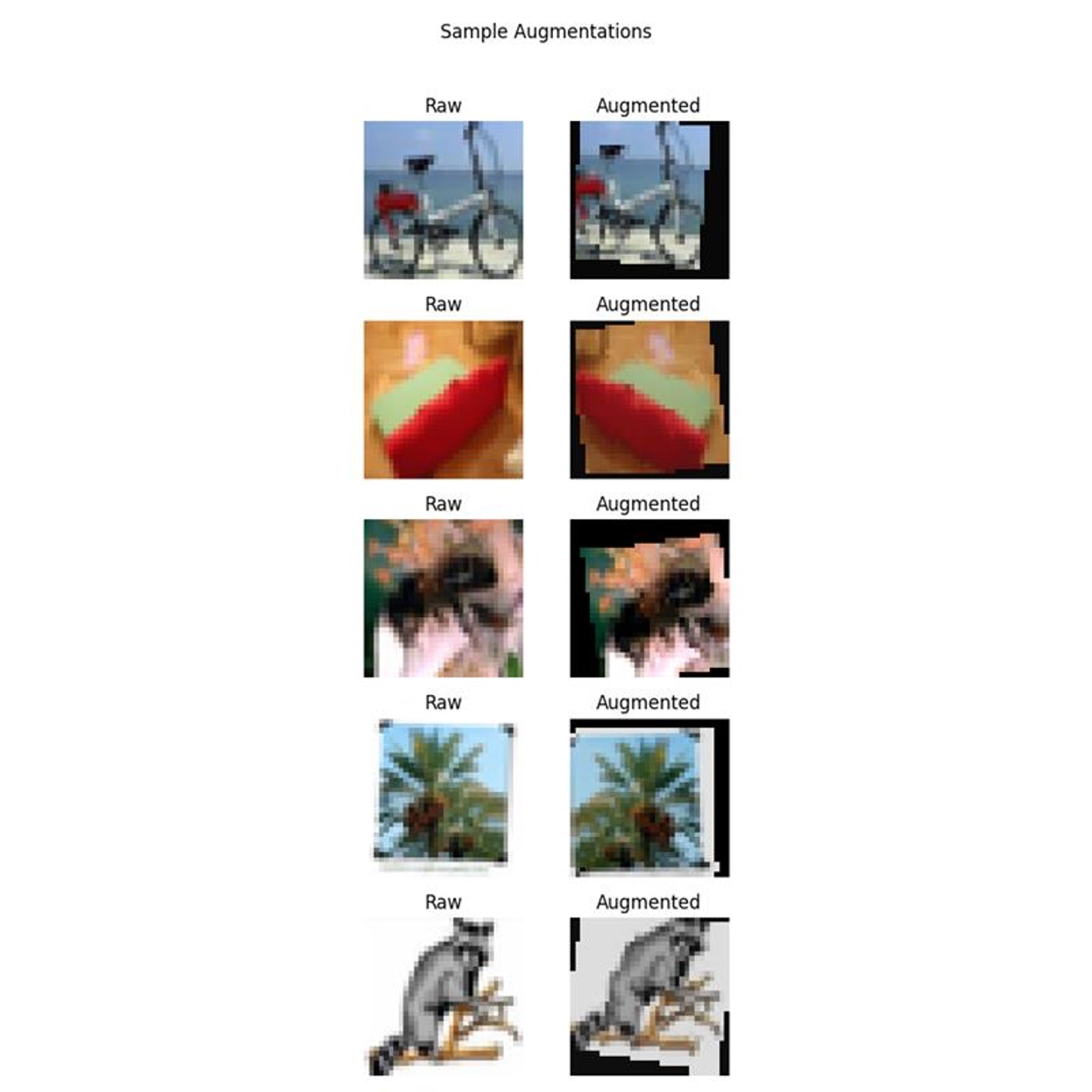}
 \caption{Sample augmentations used in both T1 and T2. Left: raw images. Right:
 augmented versions (random crop and horizontal flip). The shortcut patch is
 injected after these transforms, ensuring its location is stable in the final
 image.}
 \label{fig:augs}
\end{figure}

All images in both T1 and T2 undergo standard augmentations (random crop and
horizontal flip). Because the shortcut is injected after these
transforms, its spatial position remains fixed, avoiding confounds where
augmentation could inadvertently move or distort the cue. Representative
examples of raw versus augmented images are shown in
Figure~\ref{fig:augs}.

\subsection{Subset configuration and baselines}
We adopt an 8+4 \emph{superclass} split: 8 superclasses in T1, followed by 4 new superclasses in T2 (2 $SC$ and 2 $NSC$). This choice is both practical and methodologically motivated. First, it reduces complexity and cost relative to a full 100-class schedule, making results easier to interpret by limiting confounding interactions among many fine-grained categories. Second, operating at the superclass level encourages learning broader, more transferable representations across multiple subclasses, aligning with ERI’s goal of distinguishing genuine transfer from cue-driven performance.

We use \emph{Scratch\_T2}—a model trained from random initialization on T2 only—as the primary baseline for ERI comparisons (Adaptation Delay, Performance Deficit, and shortcut reliance).

\subsection{Models and baselines}
All methods use a ResNet-18 backbone~\cite{DBLP:journals/corr/HeZRS15}. We evaluate five continual learning strategies: 
\begin{itemize}
    \item Na\"ive sequential fine-tuning (\textbf{SGD});
    \item online Elastic Weight Consolidation (\textbf{EWC\_on})~\cite{doi:10.1073/pnas.1611835114};
    \item Dark Experience Replay (\textbf{DER++})~\cite{buzzega2020darkexperiencegeneralcontinual};
    \item Gradient Projection Memory (\textbf{GPM})~\cite{DBLP:journals/corr/abs-2103-09762}; and
    \item Deep Generative Replay (\textbf{DGR})~\cite{DBLP:journals/corr/ShinLKK17}.
\end{itemize}

\section{Results}
\label{sec:results}

\begin{figure*}[!t]
  \centering
  \includegraphics[width=\textwidth]{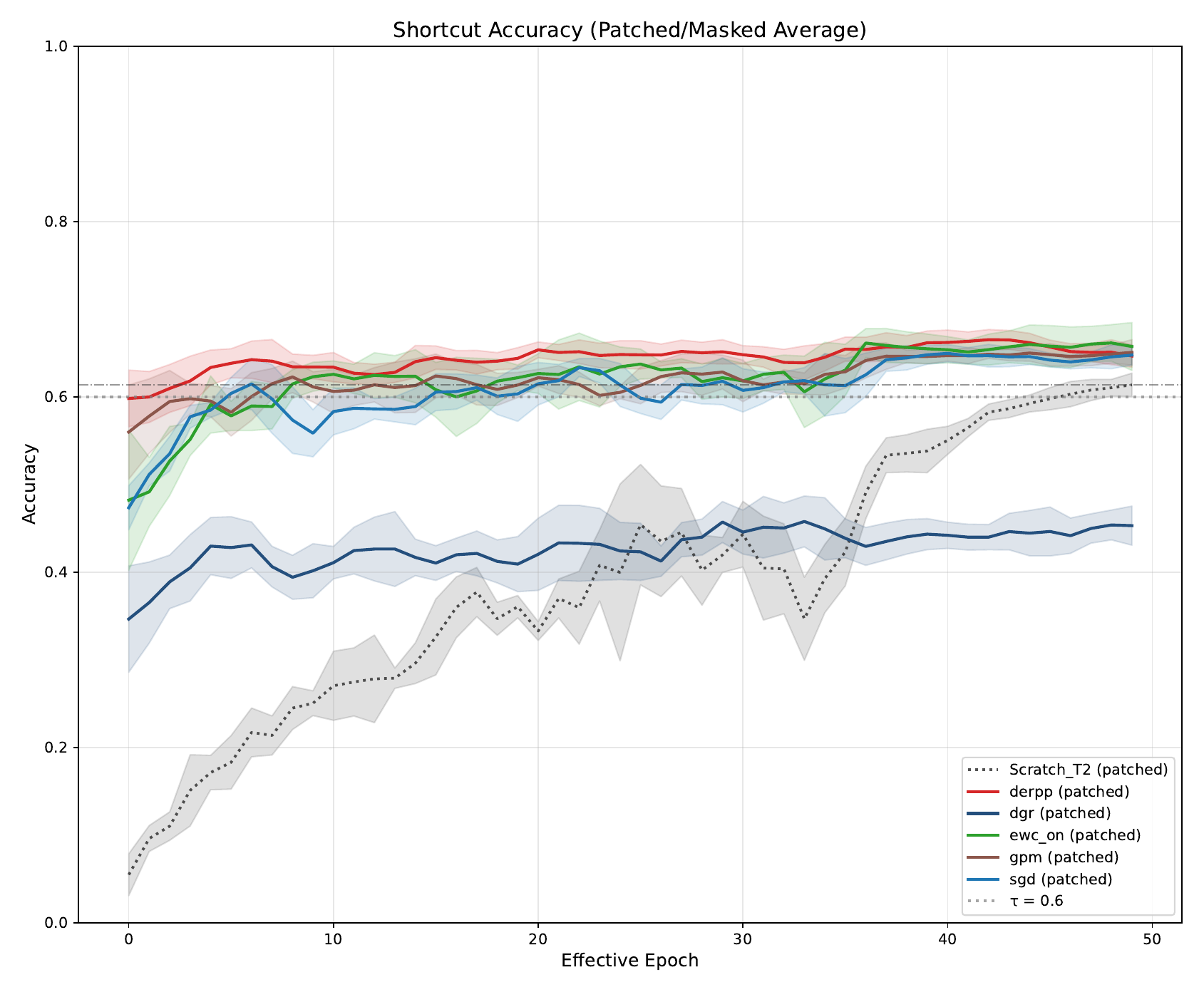}
  \caption{Panel~A: Shortcut accuracy vs.\ effective epochs on $SC$ (patched).
  AD annotations show large negative \(\mathrm{AD}\) for all continual learners
  that cross \(\tau{=}0.6\); DGR does not cross within 50 epochs.}
  \label{fig:eri_a}
\end{figure*}

\begin{figure}[!t]
  \centering
  \includegraphics[width=\columnwidth]{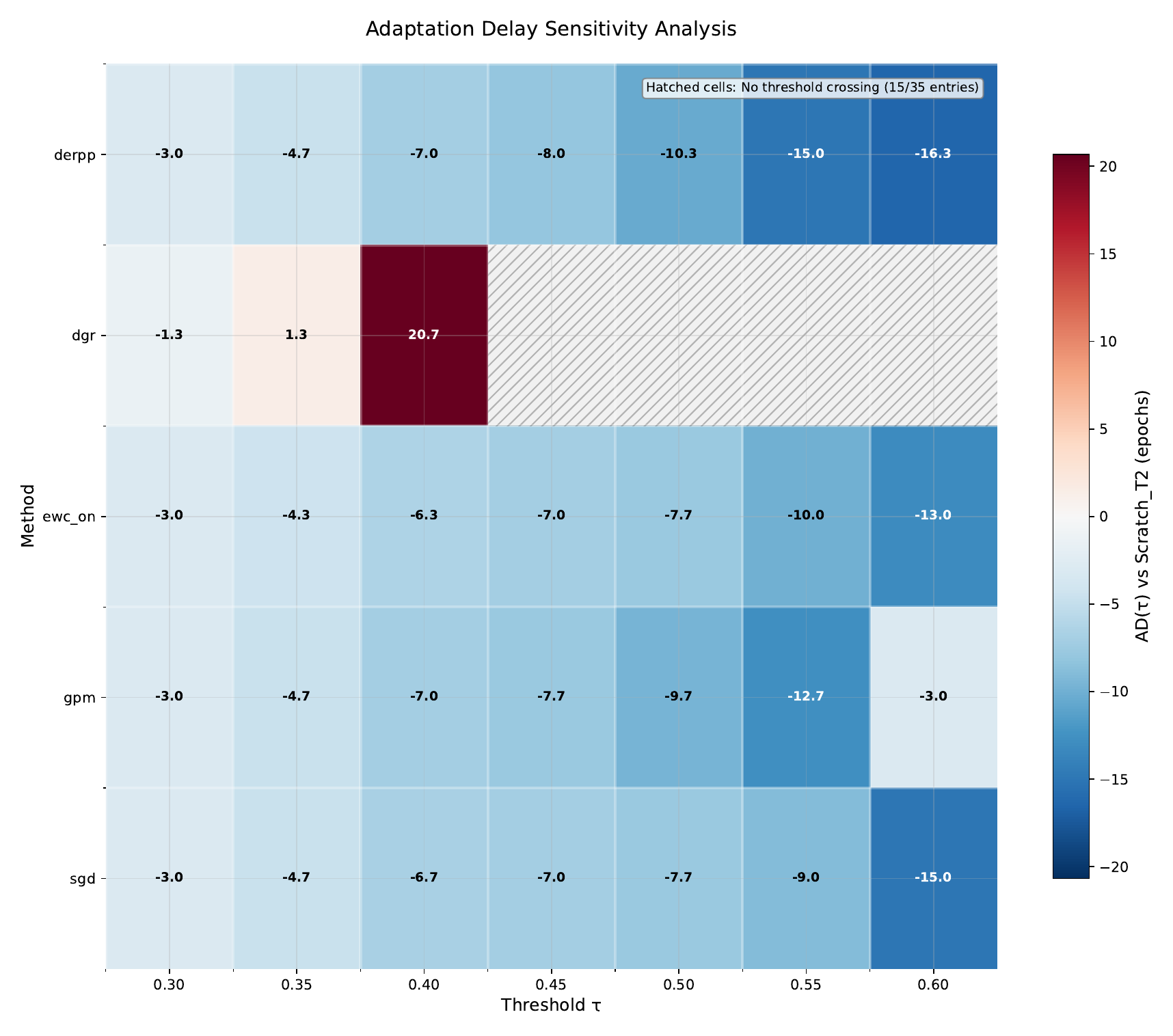}
  \caption{AD sensitivity analysis on $SC$. Negative indicates earlier adaptation than Scratch\_T2; positive indicates slower adaptation. Hatched cells denote no threshold crossing within the 50-epoch budget.}
  \label{fig:ad_heatmap}
\end{figure}

\begin{figure*}[!t]
  \centering
  \includegraphics[width=\textwidth]{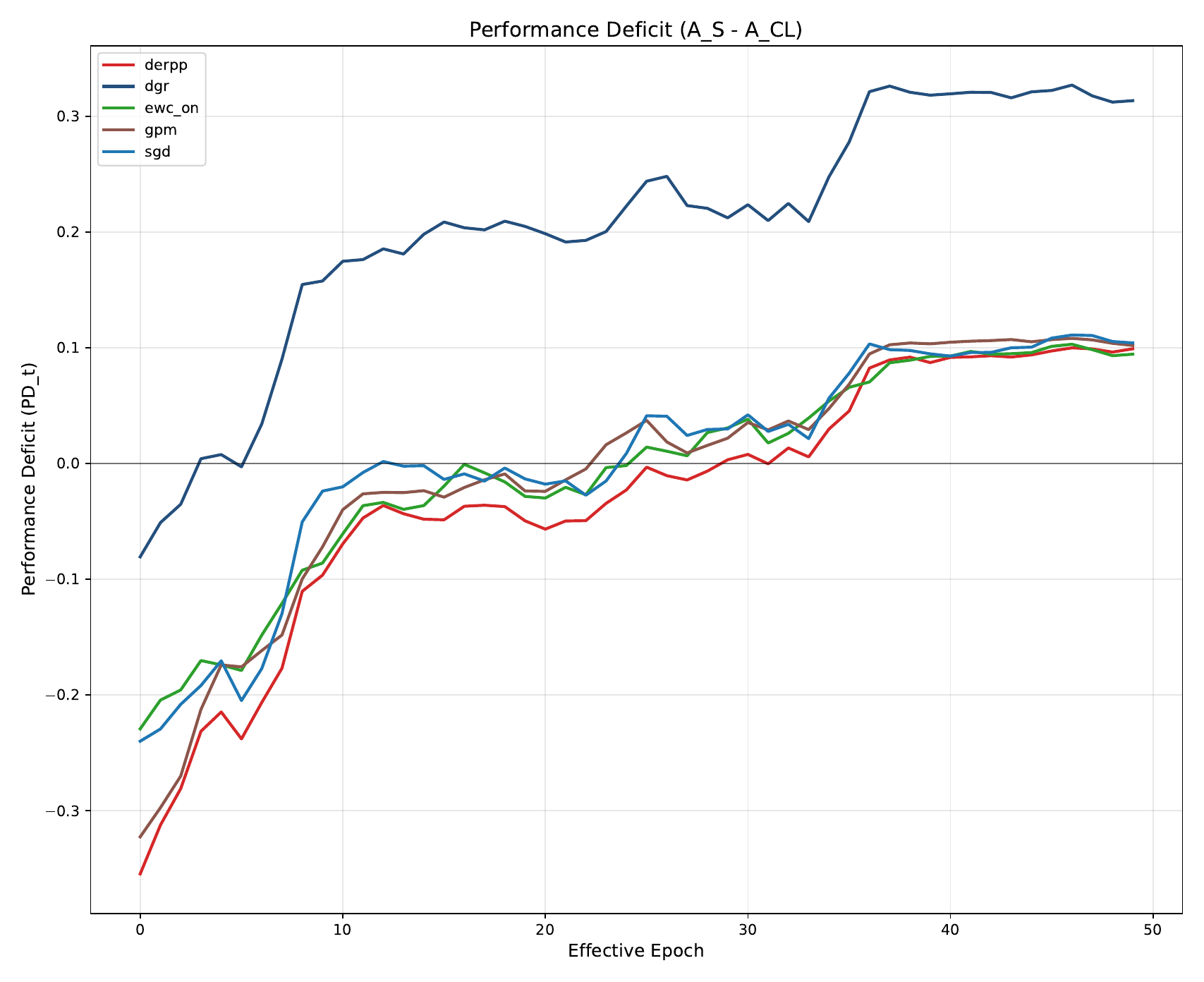}
  \caption{Panel~B: Performance Deficit over time,
  \( \mathrm{PD}_t = A_S - A_{CL}\) (SC, patched). Negative values indicate
  the CL method outperforms Scratch\_T2.}
  \label{fig:eri_b}
\end{figure*}

\begin{figure*}[!t]
  \centering
  \includegraphics[width=\textwidth]{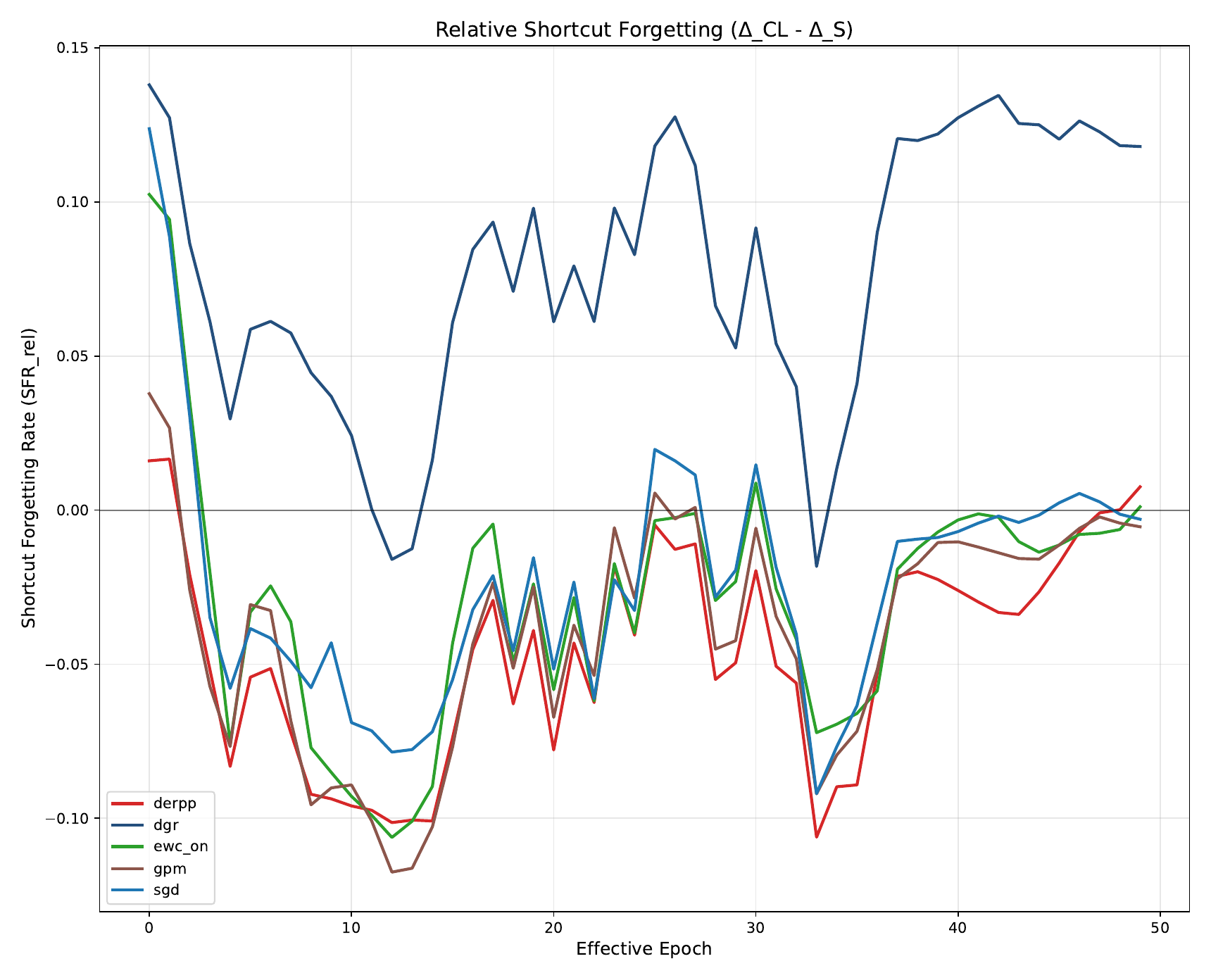}
  \caption{Panel~C: Relative shortcut sensitivity,
  \( \mathrm{SFR}_{\mathrm{rel}} = \Delta_{CL} - \Delta_S\) with
  \(\Delta = \mathrm{Acc}_{\text{patched}} - \mathrm{Acc}_{\text{masked}}\).
  In this run \(\Delta{<}0\) for all models on SC, so
  \(\mathrm{SFR}_{\mathrm{rel}}\!>\!0\) means the CL method is less hurt by the
  patch than Scratch\_T2.}
  \label{fig:eri_c}
\end{figure*}

Unless noted, all ERI quantities are computed on the designated Phase~2 shortcut superclasses (SC), use the best Phase~2 validation checkpoint for final accuracy, threshold \(\tau{=}0.6\), smoothing \(w{=}3\), and effective Phase~2 epochs for AD (to normalize replay). 
Figures~\ref{fig:eri_a}--\ref{fig:eri_c}
summarize dynamics for the three ERI facets.

To audit sensitivity of the Adaptation Delay (AD) to the choice of threshold,
we additionally evaluate \(\mathrm{AD}(\tau)\) on a grid
\(\tau \in \{0.30, 0.35, 0.40, 0.45, 0.50, 0.55, 0.60\}\) and visualize the
results as a heatmap (Fig.~\ref{fig:ad_heatmap}).

Figure~\ref{fig:eri_a} (Panel~A; \textit{Shortcut accuracy vs.\ effective epoch}) together with Fig.~\ref{fig:ad_heatmap} (\textit{AD threshold sensitivity}) shows that AD depends on the chosen threshold \(\tau\). Where crossings occur, SGD, EWC, DER++, and GPM typically exhibit moderately negative \(\mathrm{AD}(\tau)\) at moderate thresholds (earlier adaptation than Scratch\_T2), while occasional positive \(\mathrm{AD}(\tau)\) outliers indicate cases where Scratch\_T2 reaches the threshold sooner. DGR seldom reaches higher thresholds within the epoch budget, consistent with multiple no-cross regions.

Figure~\ref{fig:eri_b} (Panel~B; \textit{Performance Deficit}) shows that at convergence on $SC$ (patched) the continual learners have small but positive PD (i.e., they slightly underperform Scratch\_T2), while DGR's deficit is substantially larger. These end-point statistics match Table~\ref{tab:results_summary_new}.

Figure~\ref{fig:eri_c} (Panel~C; \textit{Relative shortcut sensitivity})
shows that Scratch\_T2 benefits slightly from the patch (\(\Delta_S{>}0\): patched \(>\) masked), whereas all CL methods are harmed by it (\(\Delta_{CL}{<}0\)). Consequently, \(\mathrm{SFR}_{\mathrm{rel}}{<}0\) for all CL methods in this run, indicating greater patch-induced harm relative to Scratch\_T2.

\begin{table*}[!t]
\centering
\caption{Summary ERI metrics (Mean \(\pm\) Standard Deviation). AD is analyzed in Fig.~\ref{fig:ad_heatmap} and not tabulated.}
\label{tab:results_summary_new}
\resizebox{\textwidth}{!}{%
\begin{tabular}{@{}lcccccc@{}}
\toprule
Strategy &
PD &
SFR\(_{\text{rel}}\) &
\(\mathcal{A}_{\text{SC}}^{\text{patch}}\) &
\(\mathcal{A}_{\text{SC}}^{\text{mask}}\) &
\(\mathcal{A}_{\text{T1}}\) &
\(\mathcal{A}_{\text{NSC}}^{\text{patch}}\) \\
\midrule
Scratch\_T2 & -- & -- &
\(0.628 \pm 0.007\) & \(0.608 \pm 0.033\) &
-- & \(0.706 \pm 0.007\) \\
derpp & \(0.027 \pm 0.008\) & \(-0.095 \pm 0.034\) &
\(0.601 \pm 0.015\) & \(0.676 \pm 0.010\) &
\(0.679 \pm 0.004\) & \(0.796 \pm 0.004\) \\
dgr & \(0.247 \pm 0.023\) & \(-0.141 \pm 0.031\) &
\(0.381 \pm 0.020\) & \(0.502 \pm 0.050\) &
\(0.682 \pm 0.010\) & \(0.730 \pm 0.007\) \\
ewc\_on & \(0.022 \pm 0.046\) & \(-0.105 \pm 0.034\) &
\(0.606 \pm 0.045\) & \(0.691 \pm 0.044\) &
\(0.691 \pm 0.007\) & \(0.804 \pm 0.006\) \\
gpm & \(0.026 \pm 0.026\) & \(-0.112 \pm 0.034\) &
\(0.602 \pm 0.026\) & \(0.694 \pm 0.024\) &
\(0.678 \pm 0.006\) & \(0.792 \pm 0.009\) \\
sgd & \(0.028 \pm 0.020\) & \(-0.115 \pm 0.028\) &
\(0.600 \pm 0.014\) & \(0.695 \pm 0.015\) &
\(0.684 \pm 0.005\) & \(0.793 \pm 0.006\) \\
\bottomrule
\end{tabular}%
}
\end{table*}

\subsection{Interpretation of the Einstellung triplet}

\paragraph{Adaptation Delay (AD)}

With \(\mathrm{AD}(\tau) = t_{\text{CL}}(\tau) - t_{\text{S}}(\tau)\), negative values indicate earlier adaptation than Scratch\_T2. The heatmap (Fig.~\ref{fig:ad_heatmap}) reveals that AD is threshold-dependent and often undefined at higher \(\tau\) due to no-cross events. Where defined, SGD, EWC, DER++, and GPM typically exhibit moderately negative \(\mathrm{AD}(\tau)\) for \(\tau \le 0.50\) (faster adaptation), while occasional positive \(\mathrm{AD}(\tau)\) outliers indicate cases where Scratch\_T2 reaches the threshold earlier. DGR rarely reaches higher thresholds within the budget.

\paragraph{Performance Deficit (PD)}
On $SC$ (patched), continual learners exhibit negative PD
(\(-0.11\pm0.04\) on average), i.e., they outperform Scratch\_T2 at convergence, reflecting the nature of continual learning methods showing a positive interference against continual tasks in general.
GPM tracks the same trend on the panel. In contrast, DGR shows positive PD
(\(0.107\pm0.039\)), underperforming Scratch\_T2. On $NSC$, all continual methods
outperform Scratch\_T2, including DGR (Table~\ref{tab:results_summary_new}),
so DGR’s deficit is specific to the $SC$ distribution.

\paragraph{Relative Shortcut Sensitivity (SFR\(_{\mathrm{rel}}\))}
In this run, masking the magenta patch \emph{improves} accuracy for all
models (\(\Delta{<}0\)), implying the patch behaves as a distractor rather
than a helpful cue. Under this sign convention, negative
\(\mathrm{SFR}_{\mathrm{rel}} = \Delta_{CL} - \Delta_S\) means the CL method
is more hurt by the presence of the patch than Scratch\_T2. We observe negative \(\mathrm{SFR}_{\mathrm{rel}}\) for DER++, EWC, GPM, and SGD, with
the largest magnitude for DGR (Table~\ref{tab:results_summary_new}).

\subsection{Rigidity versus shortcut sensitivity}
Compared to Scratch\_T2 (which lacks T1 pretraining), the continual learners
appear more \emph{rigid} with respect to the injected artifact: they adapt
rapidly on $SC$ (negative AD), achieve higher patched accuracy (negative PD), and
are less harmed by the presence of the patch than Scratch\_T2
(small positive \(\mathrm{SFR}_{\mathrm{rel}}\) under \(\Delta{<}0\)). This
pattern indicates that prior knowledge and/or constraints from continual
training reduce sensitivity to the patch, which here functions as a distractor.

DGR is the exception: it performs poorly on T2 SC, never reaches \(\tau\), and
has the highest composite ERI. Since all methods (including DGR) have strong
T1 performance, the most logical explanation is overly general
generated memory that constrains adaptation---i.e., rigidity that limits
the discovery of T2-specific discriminative features, yielding inferior SC
learning relative to GPM and DER++ despite comparable or better retention.

\section{Discussion}

The empirical pattern in Section~V—negative \(\mathrm{AD}\), small positive \(\mathrm{PD}\) on Phase~2 shortcut superclasses (SC), and negative \(\mathrm{SFR}_{\mathrm{rel}}\)—does not indicate \emph{Einstellung}-like rigidity. Continual learners reach moderate accuracy thresholds faster than Scratch\_T2 but achieve slightly lower final accuracy on the patched $SC$ set; and they improve when the patch is masked, while Scratch\_T2 relies on the patch slightly. In other words, ERI detects the absence of shortcut reuse in this run and identifies the patch as a distractor for continual learners. At the same time, strong retention on T1 and superior performance on the $NSC$ set showcase positive interference from prior training.  

A mechanistic reading helps explain these outcomes. Na\"ive fine-tuning, EWC, and replay each bias adaptation toward reusing Phase~1 features to different degrees. In our setup, that bias appears to have reduced susceptibility to the shortcut: continual learners did not latch onto the more optimal magenta patch and, in fact, performed much better when it was removed, whereas Scratch\_T2 showed mild shortcut use. One plausible explanation is that Phase~1 representations and anti-forgetting constraints nudged optimization toward more distributed, semantic cues, at the cost of slightly lower accuracy when the patch is present. By contrast, DGR underperformed on T2 and often failed to reach moderate thresholds despite strong T1 retention, suggesting that overly general generated memories over-regularized the feature space and hindered discovery of T2-specific signals—an instance of rigidity that limited adaptation.  

These findings refine how to read ERI in practice. ERI’s red-flag pattern \((\mathrm{AD}\!\ll\!0,\ \mathrm{PD}\!\le\!0,\ \mathrm{SFR}_{\mathrm{rel}}\!>\!0)\) suggests shortcut-induced rigidity. By contrast, the benign-avoidance pattern we observe here \((\mathrm{AD}\!<\!0,\ \mathrm{PD}\!>\!0,\ \mathrm{SFR}_{\mathrm{rel}}\!<\!0)\) indicates faster adaptation without shortcut reuse and that the cue functions as a distractor for the continual learners. A minimal screening workflow is to (i) fix a threshold \(\tau\) and smoothing window \(w\), compute \(\mathrm{AD}\) on patched data in effective Phase~2 epochs, (ii) compare final patched accuracies via \(\mathrm{PD}\), and (iii) estimate cue reliance via masking to obtain \(\mathrm{SFR}_{\mathrm{rel}}\). If ERI flags a red-flag triplet, additional probes—such as counterfactual tests, calibration under masking, and representational drift analyses—can corroborate shortcut use. When ERI shows the benign-avoidance triplet, it highlights reduced shortcut susceptibility and can still motivate auditing for possible distractor effects or optimization interactions. Although our study uses a synthetic cue, the same logic applies when cues arise naturally (e.g., acquisition artifacts, backgrounds, overlays) \cite{Hill2024TheRO,Hermann2023OnTF, Niu2022RoadblocksFT}.

\subsection{Operationalizing ERI when cues are unknown}
ERI’s SFR component assumes access to an intervention that neutralizes the suspected cue while preserving other content. When cues are unknown, we approximate this with lightweight proxies:
(i) Occlusion sensitivity: slide a small occluder over the image to obtain an importance map; define a budgeted mask (e.g., top 2–5\% most influential pixels) and compute \(\Delta_M\) under that mask. 
(ii) Attribution-guided deletion: aggregate attributions (e.g., gradient-based or perturbation-based) across validation images and derive a class-conditional deletion map; apply a fixed-area deletion at test time to estimate \(\Delta_M\).
(iii) Counterfactual replacements: replace backgrounds or color channels with class-agnostic content (e.g., shuffled backgrounds or inpainted regions) to remove broad classes of potential cues.
To reduce false positives, include controls: (a) apply the same masks to non-shortcut classes ($NSC$), (b) use random-location masks of equal area, and (c) vary mask color or inpainting to detect mask-induced artifacts. These proxies are imperfect and can bias \(\Delta_M\); we therefore interpret \(\mathrm{SFR}_{\mathrm{rel}}\) under unknown cues as a conservative indicator and report mask area and construction method alongside ERI.

\subsection{Limitations}
Our shortcut is synthetic and high-contrast, which may overstate effects relative to subtle, naturally occurring artifacts. We evaluate a single backbone (ResNet-18), modest task sizes, and a fixed patch position; the magnitude of ERI signals may vary with architecture, training budget, patch salience, and location. Our study uses a two-stage CIFAR-100 protocol. While this setting isolates the
shortcut effects we target, this paper doesn't cover larger-scale vision
(e.g., ImageNet subsets), domains with different nuisance factors (e.g., medical
images), and non-vision modalities (e.g., text). Extending ERI to such settings is important future work.

We report results over four seeds; larger runs would refine uncertainty estimates. Additionally, more definitive assessments should include randomized patch placement, varying salience, larger backbones (e.g., Vision Transformers), and naturally occurring artifacts; finally, extending ERI to non-vision streams (e.g., text or multimodal) will test its generality beyond image cues.

\subsection{Scope across datasets and modalities}
Conceptually, ERI requires specifying or approximating a masking operator that removes the suspected cue while leaving other content unchanged. Where cues are unknown, proxy interventions (e.g., targeted erasure guided by saliency or generative counterfactuals) could be used, but may introduce their own biases; we leave this to future work.

\section{Conclusion}
We introduced the Einstellung Rigidity Index (ERI), a compact diagnostic that disentangles genuine transfer from cue-driven performance in continual learning using three facets: \(\mathrm{AD}\), \(\mathrm{PD}\), and \(\mathrm{SFR}_{\mathrm{rel}}\). On a two-phase CIFAR-100 protocol with a controlled cue, ERI revealed faster adaptation (negative AD) with slightly lower final accuracy on patched shortcut classes (positive PD) and greater sensitivity to the injected patch (negative \(\mathrm{SFR}_{\mathrm{rel}}\)) for several CL methods—indicating that, in this configuration, the cue acts as a distractor rather than a shortcut. We clarified how to interpret \(\mathrm{SFR}_{\mathrm{rel}}\) by cue sign and provided a magnitude-based variant for cue-harmful regimes. 

While scope and budgets limit generality, ERI proves practical as a screening tool: computed alongside standard CL metrics, it improves interpretability and helps identify when adaptation proceeds for the right reasons. Future work will test ERI under naturally occurring artifacts and broader architectures, and will investigate how to turn ERI-driven feedback into effective training interventions.

\section*{Acknowledgment}
We thank Abdullah Al Forhad and the Texas Academy of Mathematics and Science for their support.

\bibliographystyle{IEEEtran}
\bibliography{references}

\begin{thebibliography}{10}
\providecommand{\url}[1]{#1}
\csname url@samestyle\endcsname
\providecommand{\newblock}{\relax}
\providecommand{\bibinfo}[2]{#2}
\providecommand{\BIBentrySTDinterwordspacing}{\spaceskip=0pt\relax}
\providecommand{\BIBentryALTinterwordstretchfactor}{4}
\providecommand{\BIBentryALTinterwordspacing}{\spaceskip=\fontdimen2\font plus
\BIBentryALTinterwordstretchfactor\fontdimen3\font minus \fontdimen4\font\relax}
\providecommand{\BIBforeignlanguage}[2]{{%
\expandafter\ifx\csname l@#1\endcsname\relax
\typeout{** WARNING: IEEEtran.bst: No hyphenation pattern has been}%
\typeout{** loaded for the language `#1'. Using the pattern for}%
\typeout{** the default language instead.}%
\else
\language=\csname l@#1\endcsname
\fi
#2}}
\providecommand{\BIBdecl}{\relax}
\BIBdecl

\bibitem{doi:10.1073/pnas.1611835114}
\BIBentryALTinterwordspacing
J.~Kirkpatrick, R.~Pascanu, N.~Rabinowitz, J.~Veness, G.~Desjardins, A.~A. Rusu, K.~Milan, J.~Quan, T.~Ramalho, A.~Grabska-Barwinska, D.~Hassabis, C.~Clopath, D.~Kumaran, and R.~Hadsell, ``Overcoming catastrophic forgetting in neural networks,'' \emph{Proceedings of the National Academy of Sciences}, vol. 114, no.~13, pp. 3521--3526, 2017. [Online]. Available: \url{https://www.pnas.org/doi/abs/10.1073/pnas.1611835114}
\BIBentrySTDinterwordspacing

\bibitem{vandeven2024continuallearningcatastrophicforgetting}
\BIBentryALTinterwordspacing
G.~M. van~de Ven, N.~Soures, and D.~Kudithipudi, ``Continual learning and catastrophic forgetting,'' 2024. [Online]. Available: \url{https://arxiv.org/abs/2403.05175}
\BIBentrySTDinterwordspacing

\bibitem{NEURIPS2019_fa7cdfad}
\BIBentryALTinterwordspacing
D.~Rolnick, A.~Ahuja, J.~Schwarz, T.~Lillicrap, and G.~Wayne, ``Experience replay for continual learning,'' in \emph{Advances in Neural Information Processing Systems}, H.~Wallach, H.~Larochelle, A.~Beygelzimer, F.~d\textquotesingle Alch\'{e}-Buc, E.~Fox, and R.~Garnett, Eds., vol.~32.\hskip 1em plus 0.5em minus 0.4em\relax Curran Associates, Inc., 2019. [Online]. Available: \url{https://proceedings.neurips.cc/paper_files/paper/2019/file/fa7cdfad1a5aaf8370ebeda47a1ff1c3-Paper.pdf}
\BIBentrySTDinterwordspacing

\bibitem{chaudhry2018efficient}
\BIBentryALTinterwordspacing
A.~Chaudhry, M.~Ranzato, M.~Rohrbach, and M.~Elhoseiny, ``Efficient lifelong learning with a-{GEM},'' in \emph{International Conference on Learning Representations}, 2019. [Online]. Available: \url{https://openreview.net/forum?id=Hkf2_sC5FX}
\BIBentrySTDinterwordspacing

\bibitem{buzzega2020darkexperiencegeneralcontinual}
\BIBentryALTinterwordspacing
P.~Buzzega, M.~Boschini, A.~Porrello, D.~Abati, and S.~Calderara, ``Dark experience for general continual learning: a strong, simple baseline,'' 2020. [Online]. Available: \url{https://arxiv.org/abs/2004.07211}
\BIBentrySTDinterwordspacing

\bibitem{pmlr-v70-zenke17a}
\BIBentryALTinterwordspacing
F.~Zenke, B.~Poole, and S.~Ganguli, ``Continual learning through synaptic intelligence,'' in \emph{Proceedings of the 34th International Conference on Machine Learning}, ser. Proceedings of Machine Learning Research, D.~Precup and Y.~W. Teh, Eds., vol.~70.\hskip 1em plus 0.5em minus 0.4em\relax PMLR, 06--11 Aug 2017, pp. 3987--3995. [Online]. Available: \url{https://proceedings.mlr.press/v70/zenke17a.html}
\BIBentrySTDinterwordspacing

\bibitem{rusu2022progressiveneuralnetworks}
\BIBentryALTinterwordspacing
A.~A. Rusu, N.~C. Rabinowitz, G.~Desjardins, H.~Soyer, J.~Kirkpatrick, K.~Kavukcuoglu, R.~Pascanu, and R.~Hadsell, ``Progressive neural networks,'' 2022. [Online]. Available: \url{https://arxiv.org/abs/1606.04671}
\BIBentrySTDinterwordspacing

\bibitem{yoon2018lifelonglearningdynamicallyexpandable}
\BIBentryALTinterwordspacing
J.~Yoon, E.~Yang, J.~Lee, and S.~J. Hwang, ``Lifelong learning with dynamically expandable networks,'' 2018. [Online]. Available: \url{https://arxiv.org/abs/1708.01547}
\BIBentrySTDinterwordspacing

\bibitem{Wang_2021_CVPR}
S.~Wang, X.~Li, J.~Sun, and Z.~Xu, ``Training networks in null space of feature covariance for continual learning,'' in \emph{Proceedings of the IEEE/CVF Conference on Computer Vision and Pattern Recognition (CVPR)}, June 2021, pp. 184--193.

\bibitem{DBLP:journals/corr/ShinLKK17}
\BIBentryALTinterwordspacing
H.~Shin, J.~K. Lee, J.~Kim, and J.~Kim, ``Continual learning with deep generative replay,'' \emph{CoRR}, vol. abs/1705.08690, 2017. [Online]. Available: \url{http://arxiv.org/abs/1705.08690}
\BIBentrySTDinterwordspacing

\bibitem{DBLP:journals/corr/abs-2103-09762}
\BIBentryALTinterwordspacing
G.~Saha, I.~Garg, and K.~Roy, ``Gradient projection memory for continual learning,'' \emph{CoRR}, vol. abs/2103.09762, 2021. [Online]. Available: \url{https://arxiv.org/abs/2103.09762}
\BIBentrySTDinterwordspacing

\bibitem{Binz2021ReconstructingTE}
\BIBentryALTinterwordspacing
M.~Binz and E.~Schulz, ``Reconstructing the einstellung effect,'' \emph{Computational Brain \& Behavior}, vol.~6, pp. 526--542, 2021. [Online]. Available: \url{https://doi.org/10.1007/s42113-022-00161-2}
\BIBentrySTDinterwordspacing

\bibitem{Kong2022BalancingSA}
Y.~Kong, L.~Liu, Z.~Wang, and D.~Tao, \emph{Balancing Stability and Plasticity Through Advanced Null Space in Continual Learning}, 11 2022, pp. 219--236.

\bibitem{Geirhos_2020}
\BIBentryALTinterwordspacing
R.~Geirhos, J.-H. Jacobsen, C.~Michaelis, R.~Zemel, W.~Brendel, M.~Bethge, and F.~A. Wichmann, ``Shortcut learning in deep neural networks,'' \emph{Nature Machine Intelligence}, vol.~2, no.~11, p. 665–673, Nov. 2020. [Online]. Available: \url{http://dx.doi.org/10.1038/s42256-020-00257-z}
\BIBentrySTDinterwordspacing

\bibitem{Hermann2023OnTF}
\BIBentryALTinterwordspacing
K.~Hermann, H.~Mobahi, T.~Fel, and M.~C. Mozer, ``On the foundations of shortcut learning,'' \emph{ArXiv}, vol. abs/2310.16228, 2023. [Online]. Available: \url{https://arxiv.org/pdf/2310.16228}
\BIBentrySTDinterwordspacing

\bibitem{Murali2023ShortcutLT}
N.~Murali, A.~Puli, K.~Yu, R.~Ranganath, and K.~Batmanghelich, ``Shortcut learning through the lens of early training dynamics,'' 02 2023.

\bibitem{Steinmann2024navigating}
D.~Steinmann, F.~Divo, M.~Kraus, A.~Wüst, L.~Struppek, F.~Friedrich, and K.~Kersting, ``Navigating shortcuts, spurious correlations, and confounders: From origins via detection to mitigation,'' 12 2024.

\bibitem{Hill2024TheRO}
\BIBentryALTinterwordspacing
B.~G. Hill, F.~L. Koback, and P.~L. Schilling, ``The risk of shortcutting in deep learning algorithms for medical imaging research,'' \emph{Scientific Reports}, vol.~14, 2024. [Online]. Available: \url{https://doi.org/10.1038/s41598-024-79838-6}
\BIBentrySTDinterwordspacing

\bibitem{Bassi2024improving}
P.~Bassi, S.~Dertkigil, and A.~Cavalli, ``Improving deep neural network generalization and robustness to background bias via layer-wise relevance propagation optimization,'' \emph{Nature Communications}, vol.~15, 01 2024.

\bibitem{shim2023constructperfectworsethancoinflipspoofing}
\BIBentryALTinterwordspacing
H.~jin Shim, R.~G. Hautamäki, M.~Sahidullah, and T.~Kinnunen, ``How to construct perfect and worse-than-coin-flip spoofing countermeasures: A word of warning on shortcut learning,'' in \emph{Proc. Interspeech 2023}, 2023, pp. 785--789. [Online]. Available: \url{https://arxiv.org/abs/2306.00044}
\BIBentrySTDinterwordspacing

\bibitem{Adnan2022MonitoringSL}
\BIBentryALTinterwordspacing
M.~Adnan, Y.~Ioannou, C.-Y. Tsai, A.~Galloway, H.~R. Tizhoosh, and G.~W. Taylor, ``Monitoring shortcut learning using mutual information,'' \emph{ArXiv}, vol. abs/2206.13034, 2022. [Online]. Available: \url{https://arxiv.org/pdf/2206.13034}
\BIBentrySTDinterwordspacing

\bibitem{Fay2023AvoidingSB}
\BIBentryALTinterwordspacing
L.~Fay, E.~Cobos, B.~Yang, S.~Gatidis, and T.~Kuestner, ``Avoiding shortcut-learning by mutual information minimization in deep learning-based image processing,'' \emph{IEEE Access}, vol.~11, pp. 64\,070--64\,086, 2023. [Online]. Available: \url{https://ieeexplore.ieee.org/stamp/stamp.jsp?tp=&arnumber=10162210}
\BIBentrySTDinterwordspacing

\bibitem{Robinson2022DeepLM}
\BIBentryALTinterwordspacing
C.~G. Robinson, A.~Trivedi, M.~Blazes, A.~Ortiz, J.~Desbiens, S.~Gupta, R.~Dodhia, P.~K. Bhatraju, W.~C. Liles, A.~Lee, J.~Kalpathy-Cramer, and J.~M.~L. Ferres, ``Deep learning models for covid-19 chest x-ray classification: Preventing shortcut learning using feature disentanglement,'' \emph{PLoS ONE}, vol.~17, 2022. [Online]. Available: \url{https://doi.org/10.1101/2021.02.11.20196766}
\BIBentrySTDinterwordspacing

\bibitem{murali2023distributionshiftspuriousfeatures}
\BIBentryALTinterwordspacing
N.~Murali, A.~Puli, K.~Yu, R.~Ranganath, and K.~Batmanghelich, ``Beyond distribution shift: Spurious features through the lens of training dynamics,'' \emph{Transactions on Machine Learning Research}, vol. 2023, October 2023. [Online]. Available: \url{https://arxiv.org/abs/2302.09344}
\BIBentrySTDinterwordspacing

\bibitem{vandeVen2024ContinualLA}
\BIBentryALTinterwordspacing
G.~M. van~de Ven, N.~Soures, and D.~Kudithipudi, ``Continual learning and catastrophic forgetting,'' \emph{ArXiv}, vol. abs/2403.05175, 2024. [Online]. Available: \url{https://arxiv.org/pdf/2403.05175}
\BIBentrySTDinterwordspacing

\bibitem{Ashley2021DoesTA}
\BIBentryALTinterwordspacing
D.~R. Ashley, S.~Ghiassian, and R.~S. Sutton, ``Does the adam optimizer exacerbate catastrophic forgetting?'' 2021. [Online]. Available: \url{https://arxiv.org/pdf/2102.07686}
\BIBentrySTDinterwordspacing

\bibitem{Marconato2022CatastrophicFI}
\BIBentryALTinterwordspacing
E.~Marconato, G.~Bontempo, S.~Teso, E.~Ficarra, S.~Calderara, and A.~Passerini, ``Catastrophic forgetting in continual concept bottleneck models,'' in \emph{ICIAP Workshops}, 2022. [Online]. Available: \url{https://doi.org/10.1007/978-3-031-13324-4_46}
\BIBentrySTDinterwordspacing

\bibitem{Wang2023WhereTF}
\BIBentryALTinterwordspacing
H.~Wang, Q.~Wu, H.~Li, and F.~Meng, ``Where to forget: A new attention stability metric for continual learning evaluation,'' in \emph{International Forum on Digital TV and Wireless Multimedia Communication}, 2023. [Online]. Available: \url{https://doi.org/10.1007/978-981-97-3623-2_10}
\BIBentrySTDinterwordspacing

\bibitem{Sun2024RevivingDM}
\BIBentryALTinterwordspacing
H.~Sun and Y.~Gao, ``Reviving dormant memories: Investigating catastrophic forgetting in language models through rationale-guidance difficulty,'' \emph{ArXiv}, vol. abs/2411.11932, 2024. [Online]. Available: \url{https://arxiv.org/pdf/2411.11932}
\BIBentrySTDinterwordspacing

\bibitem{hammoud2023rapidadaptationonlinecontinual}
\BIBentryALTinterwordspacing
H.~A. A.~K. Hammoud, A.~Prabhu, S.-N. Lim, P.~H.~S. Torr, A.~Bibi, and B.~Ghanem, ``Rapid adaptation in online continual learning: Are we evaluating it right?'' in \emph{ICCV 2023}, 2023, pp. 18\,852--18\,861. [Online]. Available: \url{https://arxiv.org/abs/2305.09275}
\BIBentrySTDinterwordspacing

\bibitem{BAKER2023274}
\BIBentryALTinterwordspacing
M.~M. Baker, A.~New, M.~Aguilar-Simon, Z.~Al-Halah, S.~M. Arnold, E.~Ben-Iwhiwhu, A.~P. Brna, E.~Brooks, R.~C. Brown, Z.~Daniels, A.~Daram, F.~Delattre, R.~Dellana, E.~Eaton, H.~Fu, K.~Grauman, J.~Hostetler, S.~Iqbal, C.~Kent, N.~Ketz, S.~Kolouri, G.~Konidaris, D.~Kudithipudi, E.~Learned-Miller, S.~Lee, M.~L. Littman, S.~Madireddy, J.~A. Mendez, E.~Q. Nguyen, C.~Piatko, P.~K. Pilly, A.~Raghavan, A.~Rahman, S.~K. Ramakrishnan, N.~Ratzlaff, A.~Soltoggio, P.~Stone, I.~Sur, Z.~Tang, S.~Tiwari, K.~Vedder, F.~Wang, Z.~Xu, A.~Yanguas-Gil, H.~Yedidsion, S.~Yu, and G.~K. Vallabha, ``A domain-agnostic approach for characterization of lifelong learning systems,'' \emph{Neural Networks}, vol. 160, pp. 274--296, 2023. [Online]. Available: \url{https://www.sciencedirect.com/science/article/pii/S0893608023000072}
\BIBentrySTDinterwordspacing

\bibitem{salman2022doesbiastransfertransfer}
\BIBentryALTinterwordspacing
H.~Salman, S.~Jain, A.~Ilyas, L.~Engstrom, E.~Wong, and A.~Madry, ``When does bias transfer in transfer learning?'' 2022. [Online]. Available: \url{https://arxiv.org/abs/2207.02842}
\BIBentrySTDinterwordspacing

\bibitem{salmani2024transferlearningbias}
P.~Salmani and P.~Lewis, \emph{Transfer Learning Can Introduce Bias}, 10 2024.

\bibitem{10.1162/opmi_a_00158}
\BIBentryALTinterwordspacing
A.~Székely, B.~Török, M.~Kiss, K.~Janacsek, D.~Németh, and G.~Orbán, ``Identifying transfer learning in the reshaping of inductive biases,'' \emph{Open Mind}, vol.~8, pp. 1107--1128, 09 2024. [Online]. Available: \url{https://doi.org/10.1162/opmi_a_00158}
\BIBentrySTDinterwordspacing

\bibitem{Krizhevsky2009LearningML}
\BIBentryALTinterwordspacing
A.~Krizhevsky, ``Learning multiple layers of features from tiny images,'' 2009. [Online]. Available: \url{https://www.cs.toronto.edu/~kriz/learning-features-2009-TR.pdf}
\BIBentrySTDinterwordspacing

\bibitem{DBLP:journals/corr/HeZRS15}
\BIBentryALTinterwordspacing
K.~He, X.~Zhang, S.~Ren, and J.~Sun, ``Deep residual learning for image recognition,'' \emph{CoRR}, vol. abs/1512.03385, 2015. [Online]. Available: \url{http://arxiv.org/abs/1512.03385}
\BIBentrySTDinterwordspacing

\bibitem{Niu2022RoadblocksFT}
\BIBentryALTinterwordspacing
H.~Niu, H.~Li, F.~Zhao, and B.~Li, ``Roadblocks for temporarily disabling shortcuts and learning new knowledge,'' in \emph{Advances in Neural Information Processing Systems}, S.~Koyejo, S.~Mohamed, A.~Agarwal, D.~Belgrave, K.~Cho, and A.~Oh, Eds., vol.~35.\hskip 1em plus 0.5em minus 0.4em\relax Curran Associates, Inc., 2022, pp. 29\,064--29\,075. [Online]. Available: \url{https://proceedings.neurips.cc/paper_files/paper/2022/file/baaa7b5b5bbaadca5023e1ab909b8af5-Paper-Conference.pdf}
\BIBentrySTDinterwordspacing

\end{thebibliography}

\end{document}